\documentclass[10pt,twocolumn,letterpaper]{article}

\usepackage{cvpr}              %

\newcommand{\parsection}[1]{\noindent\textbf{#1:}}
\def\modelname{{IDSplat}}
\usepackage{multirow}
\usepackage{amssymb}
\usepackage{tabulary}
\usepackage{colortbl}
\usepackage{xcolor}
\usepackage{arydshln}
\definecolor{tabfirst}{rgb}{1, 0.7, 0.7} %
\definecolor{tabsecond}{rgb}{1, 0.85, 0.7} %
\definecolor{tabthird}{rgb}{1, 1, 0.7} %
\definecolor{gray}{rgb}{0.85, 0.85, 0.85} %

\usepackage{comment}

\definecolor{cvprblue}{rgb}{0.21,0.49,0.74}
\usepackage[pagebackref,breaklinks,colorlinks,allcolors=cvprblue]{hyperref}

\def\paperID{40508} %
\def\confName{CVPR}
\def\confYear{2026}

\title{IDSplat: Instance-Decomposed 3D Gaussian Splatting for Driving Scenes}

\author{Carl Lindström$^{\dagger,1,2}$
\quad Mahan Rafidashti$^{\dagger,1,2}$
\quad Maryam Fatemi$^{1}$ \\
\quad Lars Hammarstrand$^{2}$
\quad Martin R. Oswald$^{3}$
\quad Lennart Svensson$^{2}$
\\
\normalsize$^1$Zenseact \hspace{0.8cm} $^2$Chalmers University of Technology \hspace{0.8cm} $^3$University of Amsterdam \hspace{0.8cm}\\
{\tt\small \{firstname.lastname\}@\{zenseact.com, chalmers.se\}}
}

\newread\imgstream
\immediate\openin\imgstream=imagedata.in
\makeatletter
\def\new@kvginclip#1{}
\def\new@kvgintrim#1{}
\let\old@kvginclip\KV@Gin@clip
\let\old@kvgintrim\KV@Gin@trim
\let\oldincludegraphics\includegraphics
\providecommand{\includegraphics}{}
\renewcommand{\includegraphics}[2][]{%
  \immediate\read\imgstream to \src
  \immediate\read\imgstream to \removecrop
  \ifnum\removecrop=1
      \let\KV@Gin@clip\new@kvginclip
      \let\KV@Gin@trim\new@kvgintrim
  \fi
  \oldincludegraphics[#1]{\src}%
  \let\KV@Gin@clip\old@kvginclip
  \let\KV@Gin@trim\old@kvgintrim}
\makeatother

\begin{document}
\twocolumn[{
    \maketitle
    \vspace{-10mm}
    \begin{center}
        \captionsetup{type=figure}
        \includegraphics[width=0.99\textwidth,trim={0cm 0cm 0cm 0cm},clip]{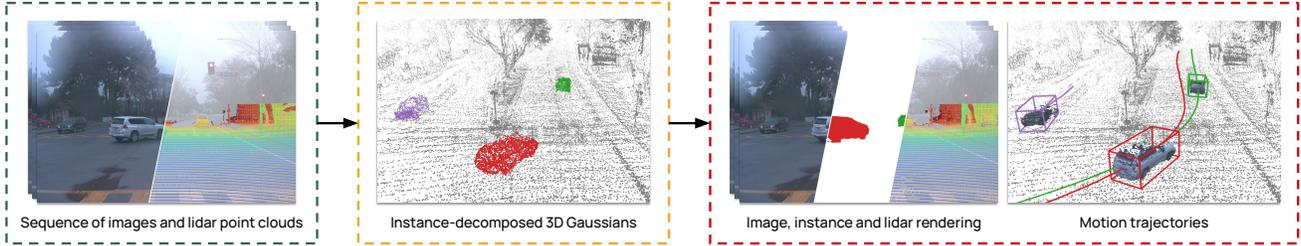}
        \captionof{figure}{{\modelname} performs self-supervised reconstruction of dynamic scenes with explicit instance-decomposition and learnable motion trajectories. {\modelname} enables high-fidelity rendering of images, instances, and lidar point clouds without the need for human annotations.
        }
        \label{fig:topfigure}
    \end{center}
}]
\maketitle
\begin{abstract}
Reconstructing dynamic driving scenes is essential for developing autonomous systems through sensor-realistic simulation. 
Although recent methods achieve high-fidelity reconstructions, they either rely on costly human annotations for object trajectories or use time-varying representations without explicit object-level decomposition, leading to intertwined static and dynamic elements that hinder scene separation.
We present {\modelname}, a self-supervised 3D Gaussian Splatting framework that reconstructs dynamic scenes with explicit instance decomposition and learnable motion trajectories, without requiring human annotations.
Our key insight is to model dynamic objects as coherent instances undergoing rigid transformations, rather than unstructured time-varying primitives.
For instance decomposition, we employ zero-shot, language-grounded video tracking anchored to 3D using lidar, and estimate consistent poses via feature correspondences.
We introduce a coordinated-turn smoothing scheme to obtain temporally and physically consistent motion trajectories, mitigating pose misalignments and tracking failures, followed by joint optimization of object poses and Gaussian parameters.
Experiments on the Waymo Open Dataset demonstrate that our method achieves competitive reconstruction quality while maintaining instance-level decomposition and generalizes across diverse sequences and view densities without retraining, making it practical for large-scale autonomous driving applications.
Code will be released.
\end{abstract}
\def\thefootnote{$\dagger$}\footnotetext{\vspace{-3mm}These authors contributed equally to this work.}\def\thefootnote{\arabic{footnote}}
\section{Introduction}
\label{sec:intro}

Reconstructing dynamic scenes has become an important cornerstone in the development of autonomous driving systems, enabling closed-loop training and testing through sensor-realistic renderings. In contrast to real-world testing, digital twins provide a scalable, low-cost, and safe means of exploring novel driving scenarios using already collected data \cite{lindstrom2024real2sim, ljungbergh2024neuroncap}.
Recent work has improved the quality, efficiency, and sensor compatibility of such reconstructions \cite{yang2023unisim, wu2024mars, tonderski2024neurad, hess2025splatad, turki2023suds, yan2024streetgs, zhou2024drivinggaussian, chen2025omnire}, but typically rely on human-annotated object trajectories and 3D bounding boxes, which are expensive and time-consuming to acquire at scale.

To address this challenge, several self-supervised approaches have emerged that aim to reconstruct dynamic scenes without human annotations \cite{yang2023emernerf, chen2026periodic, peng2025desire, xu2025ad, huang2024s3gaussian, zhang2024street}. 
Although they produce high-quality renderings, they lack explicit instance decomposition, significantly limiting their practical use, since novel scenario generation requires manipulating individual dynamic objects. %

In this paper, we address the problem of reconstructing realistic dynamic scenes without human annotations, while preserving instance-level decomposition and learning the underlying motion trajectories of individual objects. 
Reconstructing dynamic scenes using 3D Gaussian Splatting (3DGS)~\cite{kerbl20233Dgaussians} presents a significant challenge, as it violates the assumption that each 3D point maintains a fixed position and appearance across viewpoints. While prior self-supervised approaches address this by introducing time-dependent Gaussian parameters \cite{chen2026periodic,huang2024s3gaussian,peng2025desire}, we instead preserve the geometry and appearance of coherent object instances over time and optimize their rigid transformations to capture the true underlying motion. This formulation preserves instance-level decomposition and enables controllable trajectories, rather than relying on time-varying primitives whose changing visibility and appearance can lead to inconsistent scene representations. 

Obtaining instance-level decomposition of 3D Gaussians without annotated poses introduces an additional challenge. Although 3D object trackers can perform well in estimating object poses and have been used effectively in dynamic scene reconstruction \cite{tonderski2024neurad, yan2024streetgs}, they rely on human annotations for fine-tuning on the target data and are constrained by the predefined taxonomy of those annotations. To overcome this limitation, we leverage recent advances in vision models and employ a language-grounded video tracker to extract instance masks in a zero-shot manner, allowing both generalization to new datasets and new classes without retraining or additional annotations. These masks are lifted to 3D using corresponding lidar point clouds, and object poses are estimated via RANSAC using DINOv3~\cite{simeoni2025dinov3} feature correspondences. To further address pose misalignments and missing detections, we introduce an iterative robust coordinated-turn smoothing scheme that discards outliers and refines the trajectories.

Although accurate initialization of object trajectories is crucial, inaccuracies are inevitable due to missing instance masks, inaccurate lidar poses, or tracking drift over time. To mitigate these errors, we make a final refinement of the object trajectories guided by reconstruction errors during the 3D Gaussian Splatting optimization. 

We present {\modelname}, a novel 3D Gaussian Splatting framework designed to handle dynamic scenes through instance-level decomposition and joint optimization of object appearance, geometry, and motion.
We extensively evaluate the effectiveness of our method on Waymo Open Dataset \cite{sun2025waymoopendataset}, achieving state-of-the-art results across a diverse set of sequences and test protocols. In summary, our contributions are as follows:
\begin{itemize}
    \item We propose a self-supervised framework for dynamic scene reconstruction that explicitly decomposes scenes into object instances with learnable motion trajectories, enabling joint rendering, segmentation and motion tracking.
    \item We introduce a zero-shot approach for 3D instance decomposition and pose estimation, enabling generalization to new datasets and object classes without retraining or human annotations.
    \item We present simple yet effective techniques for optimizing and refining motion trajectories, combining motion modeling and photometric consistency to obtain accurate trajectories even under sparse views.
\end{itemize}
\section{Related work}
\label{sec:relatedwork}

\parsection{Annotation-based rendering for autonomous driving} 
Neural radiance fields (NeRF)~\cite{mildenhall2021nerf} inspired numerous NeRF-based methods for dynamic road scenes \cite{tonderski2024neurad, wu2024mars, yang2023unisim, rafidashti2025neuradar, chen2025omnire}. These achieve high-quality renderings and naturally extend to new sensors \cite{tonderski2024neurad, rafidashti2025neuradar}, but are sample-intensive, resulting in slow rendering speeds and limiting their scalability.

3D Gaussian Splatting (3DGS)~\cite{kerbl20233Dgaussians} provides an explicit rasterization-friendly representation, enabling orders-of-magnitude faster rendering. Automotive scene reconstruction with 3DGS has been explored in several works, including lidar-based extensions \cite{hess2025splatad, yan2024streetgs, zhou2024drivinggaussian, zhou2024hugs, chen2025g3r, khan2024autosplat, chen2025omnire}.

To model dynamic scenes, both NeRF- and 3DGS-based approaches typically rely on accurate 3D bounding boxes with temporal instance associations. These enable the scene to be decomposed into a static background and dynamic foreground components, with each dynamic object transformed according to its trajectory. To achieve high-fidelity reconstruction, state-of-the-art approaches typically rely on either human-annotated bounding boxes or predictions from high-performing 3D object trackers that have been carefully adapted to the target dataset. This reliance on curated annotations or dataset-specific trackers limits the scalability and zero-shot generalization of these methods to new datasets.

\parsection{Self-supervised dynamic scene reconstruction} 
Self-supervised approaches also separate static and dynamic regions, using either separate hash-grids \cite{yang2023emernerf, turki2023suds}, or time-varying 3DGS representations \cite{wu20244dgs, xu2024grid4d, chen2026periodic, yang2025storm, huang2024s3gaussian, peng2025desire, song2025coda, sun2025splatflow, mao2025unire}, where Gaussian attributes are allowed to change over time. Instead of bounding boxes, these methods rely on photometric and geometric consistency or foundation model outputs to guide decomposition, either via learned features or explicitly via predicted masks \cite{wu20244dgs, yang2024deform, huang2024s3gaussian, peng2025desire, xu2025ad, zhang2024street, yang2023emernerf}.
DeSiRe-GS~\cite{peng2025desire} segments dynamic regions in images using features from FiT3D~\cite{yue2025fit3d} and uses these dynamic masks to optimize time-varying Gaussians. Similarly, AD-GS~\cite{xu2025ad} relies on Grounded-SAM-2~\cite{groundedsam2} masks, but uses B-splines and trigonometric functions to enforce smoother trajectories for individual Gaussians. Other works attempt more detailed decomposition, such as CoDa-4DGS \cite{song2025coda} which enables semantic segmentation by learning per-Gaussian feature vectors. A key limitation of these models is that temporal changes are modeled at the primitive level, preventing decomposition into coherent object instances. This restricts practical applications such as novel scenario generation, auto-labeling, and simulation. %

{\modelname} addresses this by decomposing scenes into a static background and dynamic foreground instances that remain consistent across the sequence. Each instance's motion trajectory is explicitly represented, enabling reassignment, manipulation or removal. We achieve self-supervised instance decomposition using Grounded-SAM-2 for zero-shot masks and DINOv3~\cite{simeoni2025dinov3} for robust registration across time, and guide the refinement of motion trajectories through photometric and geometric consistency.
\begin{figure*}[t]
    \centering
    \includegraphics[width=1\linewidth,trim={0cm 0cm 0cm 0cm},clip]{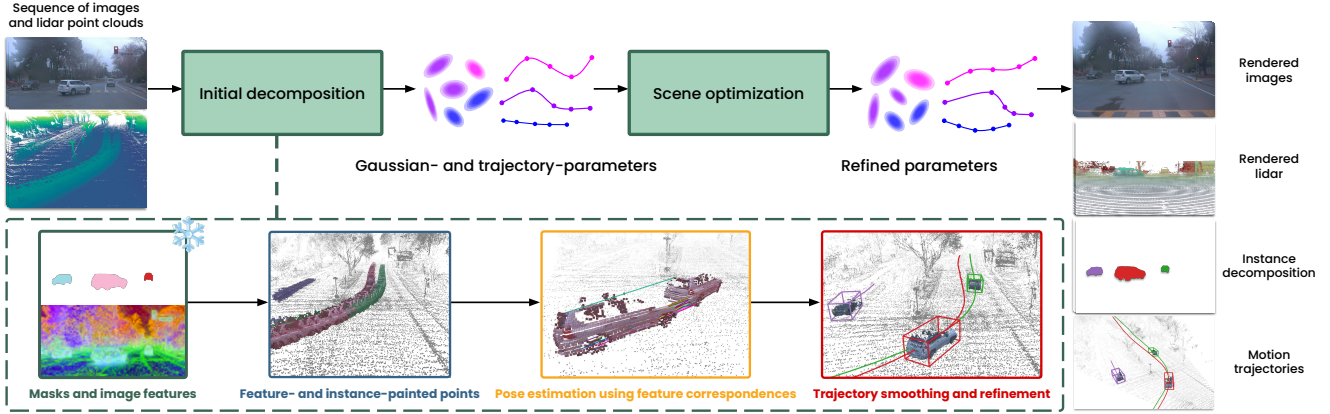}
    \caption{Overview of our method. 2D masks from Grounded-SAM-2 are lifted to 3D using corresponding lidar point clouds to initialize instances. Object poses are estimated via RANSAC using DINOv3 feature correspondences and further refined through iterative CT smoothing. Trajectories and Gaussian parameters are then optimized to render images, lidar, and instances with motion trajectories.}
    \label{fig:method_overview}
\end{figure*}
\section{Method}
\label{sec:Method}

Given a sequence of images and lidar point clouds from a driving scenario, our goal is to learn a 3D representation of the scene, including instance-decomposed dynamic objects with associated learnable motion trajectories. In the following, we describe our scene representation (\cref{sec:scene_representation}), how the representation is decomposed into separate instances (\cref{sec:instance_decomposition}), and how associated trajectories are estimated and refined (\cref{sec:point_registration,sec:trajectory_smoothing}). Finally, we describe how the complete scene is optimized (\cref{sec:scene_optimization}). See \cref{fig:method_overview} for an overview of our method.

\subsection{Scene representation}\label{sec:scene_representation}
We represent the scene by a set of translucent 3D Gaussians, parameterized with occupancy probability $o \in [0,1]$, mean $\mathbf{\mu} \in \mathbb{R}^3$, and covariance $\mathbf{\Sigma} \in \mathbb{R}^{3 \times 3}$. Together, the parameters describe the position, extent and visibility of the Gaussian. To facilitate both camera and lidar rendering, we follow SplatAD \cite{hess2025splatad} and assign each Gaussian a feature vector $\mathbf{f}^{\text{rgb}} \in \mathbb{R}^3$ to represent its base color, and another feature vector $\mathbf{f} \in \mathbb{R}^{D_{\text{f}}}$ to represent view-dependent effects and lidar properties. Further, each Gaussian also has an associated discrete ID denoted by $z \in \{0, \dots, N_{\text{ID}}\}$, determining which instance it belongs to, or whether it belongs to the static background ($z=0$). We also adopt sensor-specific embeddings to model appearance shifts.

To account for scene dynamics, we parameterize the motion of Gaussians using SE(3) poses. We assume each instance corresponds to a rigid object and transform all Gaussians associated with a given ID using the same rigid transformation. The position of Gaussian $i$ in the world at time $t$ is given as
\begin{equation}
    \mathbf{\mu}_{i,t} = \mathbf{T}_{z_i,t}\mathbf{\mu}_i
\end{equation}
where $\mathbf{\mu}_{i,t}$ denotes the Gaussian's position in world coordinates at time $t$, $\mathbf{\mu}_i$ is its position in the canonical coordinate system of instance $z_i$, and $\mathbf{T}_{z_i,t} \in \text{SE(3)}$ represents the transformation from that canonical frame to the world at time $t$. Gaussians associated with the static background are defined directly in the world coordinate system and are not transformed.

\subsection{Instance decomposition}\label{sec:instance_decomposition}
To obtain object instances, we employ Grounded-SAM-2 \cite{groundedsam2} to generate instance masks from video frames using class prompts. As our scene is represented in 3D, we lift the 2D instance information to 3D by projecting the temporally closest lidar points onto the image plane and assigning them the corresponding instance IDs. To reduce outliers from inaccurate projections, arising from line-of-sight mismatches due to camera-lidar mounting offsets or large uncorrected object motions, we apply a two-stage filtering process. We first erode the instance masks prior to projection to reduce the influence from sensor mounting offsets, and then cluster the projected points using DBSCAN \cite{ester1996density}, retaining only the largest cluster for each instance.

\subsection{Trajectory estimation}\label{sec:point_registration}
With the instance points identified, we estimate the trajectory of each object instance by registering the lidar points associated with that instance. For an instance $z$, we first define a canonical frame centered at the midpoint of the axis-aligned bounding box enclosing the points from the time step with the densest lidar observations. This defines the initial pose as
\begin{equation}
    \mathbf{T}_{z, t_{\text{init}}(z)} =  \begin{bmatrix} \mathbf{I}_{3 \times 3} & \mathbf{c}_z \\ \mathbf{0}^T & 1 \\ \end{bmatrix},
\end{equation}
where $t_{\text{init}}(z)$ denotes the time with the highest lidar density for instance $z$, and $\mathbf{c}_z$ is the center of the corresponding bounding box.
Subsequent frames are registered to the canonical frame, $\mathbf{T}_{z, t_{\text{init}}(z)}$, ordered by point density. We extract DINOv3~\cite{simeoni2025dinov3} features from image projections and establish correspondences between frames based on cosine similarity. The rigid transformation is then estimated using RANSAC, where each hypothesis is computed via the Umeyama estimator~\cite{umeyama2002least} from three randomly sampled correspondences. The pose with the largest number of structural inliers is selected if the registration is deemed successful. We consider the registration successful when the inlier ratio of the target point cloud exceeds a predefined threshold, motivated by the fact that the target (from frame $t_j$) will always be smaller than the canonical source. When a registration succeeds, the resulting pose for time $t_j$ is added to the trajectory and given by
\begin{equation}
    \mathbf{T}_{z, t_j} = \mathbf{T}_{z, t_j  \leftarrow t_{\text{init}}(z)} \mathbf{T}_{z, t_{\text{init}}(z)},
\end{equation}
where $\mathbf{T}_{z, t_j  \leftarrow t_{\text{init}}(z)}$ is the rigid transformation estimated from RANSAC, and $\mathbf{T}_{z, t_j}$ represents the pose of instance $z$ at time $t_j$. The corresponding points from frame $t_j$ are then transformed into the canonical frame using the inverse of $\mathbf{T}_{z, t_j  \leftarrow t_{\text{init}}(z)}$ and merged into the canonical point set.

\subsection{Trajectory smoothing}\label{sec:trajectory_smoothing}
While the RANSAC-based registration yields initial pose estimates between pairs of point clouds, our goal is to estimate a temporally and physically consistent object trajectory, and reduce the impact from imperfections due to registration errors or temporal gaps from missing instance masks. To address this, we refine the trajectories through an iterative coordinated-turn (CT) smoothing formulated as a pose graph optimization problem.

\parsection{Optimization formulation}
We employ GTSAM \cite{gtsam} to optimize a state vector comprising poses $\mathbf{T}_t \in \text{SE(3)}$, translational speeds $v_t \in \mathbb{R}$, and curvatures $\kappa_t \in \mathbb{R}$ for all timesteps $t$ in the sequence. The estimated poses from RANSAC are added as noisy measurement factors through Gaussian likelihoods with a Huber loss function. Since instance masks may be intermittent, measurements may only be available at a subset of timesteps. We incorporate CT motion model factors to encourage physically grounded, temporally smooth trajectory states and to reject measurement outliers. To align the axis-aligned measurements with the motion model, we also estimate a single rotation, shared across all times, that orients the local x-axis along the direction of motion. Smoothness priors on the trajectory is further added as random walk processes on speed and curvature, and additional priors are added to encourage small roll and pitch angles as well as moderate curvature values.

\parsection{Outlier rejection}
To address outliers, we apply iterative refinement of the trajectories by first performing a single optimization pass and then identifying and removing measurements whose residuals exceed a predefined threshold. The optimization is then re-run with the pruned measurement set, improving the robustness of trajectory estimates even in the presence of registration failures.

\subsection{Scene optimization}\label{sec:scene_optimization}
Our scene representation consists of a set of static Gaussians and a set of dynamic Gaussians associated to instances with corresponding trajectories. All Gaussian parameters and trajectories are optimized jointly through self-supervised reconstruction of images and lidar point clouds. We adopt the rasterization proposed in \cite{hess2025splatad} and optimize the entire model jointly using the reconstruction loss
\begin{multline}\label{eq:loss_fn}
    \mathcal{L} = \lambda_r\mathcal{L}_1 + (1-\lambda_r)\mathcal{L}_\text{SSIM} + \mathcal{L}_\text{lidar}  + \lambda_\text{MCMC}\mathcal{L}_\text{MCMC},
\end{multline}
where $\mathcal{L}_1$ and $\mathcal{L}_\text{SSIM}$ are L1 and SSIM losses on the rendered images, and $\mathcal{L}_\text{MCMC}$ denotes the opacity and scale regularization used in \cite{kheradmand2024mcmc}. $\mathcal{L}_\text{lidar}$ is the loss from lidar reconstruction and is defined as
\begin{multline}
    \mathcal{L}_\text{lidar} = \lambda_\text{depth} \mathcal{L}_\text{depth} +
    \lambda_\text{los} \mathcal{L}_\text{los} + 
    \lambda_\text{inten} \mathcal{L}_\text{inten} + \lambda_\text{raydrop}\mathcal{L}_\text{BCE},
\end{multline}
where $\mathcal{L}_\text{depth}$ and $\mathcal{L}_\text{inten}$ are L2 losses on the rendered expected lidar range and intensity, and $ \mathcal{L}_\text{los}$ is a line-of-sight loss that penalize accumulated opacity before the ground truth lidar range. $\mathcal{L}_\text{BCE}$ is a binary cross-entropy loss on predicted ray drop probability. For ease of comparison, we adopt the hyperparameters from \cite{hess2025splatad}. See \cref{sec:implementation_details} for details.

Dynamic Gaussians are initialized from the canonical point sets created during point registration (\ref{sec:point_registration}), while static Gaussians are seeded from lidar points not associated with any instance. RGB values for both sets are assigned by projecting the corresponding lidar points into the temporally closest image. Additional Gaussians are sampled randomly within the lidar range, and sampled linearly in disparity beyond observed points. Following \cite{hess2025splatad}, we employ the MCMC densification strategy from \cite{kheradmand2024mcmc}.
\section{Experiments}
\label{sec:experiments}

To thoroughly evaluate {\modelname}, we benchmark its performance on novel view synthesis (NVS) and image reconstruction using Waymo Open Dataset~\cite{sun2025waymoopendataset}. We compare against state-of-the-art self-supervised methods adapted to automotive scenes under multiple evaluation protocols, spanning a range of view densities and dynamic object categories. To assess generalization, we conduct additional experiments on PandaSet~\cite{pandaset}. We further analyze the optimized trajectories using standard tracking metrics. Finally, we perform ablation studies of individual components to understand their impact on overall performance.

\subsection{Implementation}
\begin{table}[t]
    \small
    \centering
    \setlength{\tabcolsep}{1.2pt}
    \caption{NVS results on experimental settings of AD-GS and DeSiRe-GS. Results for both baselines are obtained using their official implementation. {\modelname} outperforms the baselines for both settings. \colorbox{tabfirst}{First}, \colorbox{tabsecond}{second}, \colorbox{tabthird}{third}.}
    \resizebox{1.0\linewidth}{!}{
    \begin{tabular}{ll ccccc}
    \toprule
    &          &Anno. free   &        PSNR $\uparrow$     &       SSIM $\uparrow$     &        LPIPS $\downarrow$  &        DPSNR $\uparrow$     \\
    \midrule

    \multirow{7}{*}{\rotatebox[origin=c]{90}{\shortstack[c]{\small DeSiRe-GS setting}}}
    & MARS     &$\times$  &                                {26.61} &                                   {-} &                                    {-} &                             {22.21} \\
    & SplatAD  &$\times$  &                                {30.80} &                               {0.900} &                                {0.160} &                             {28.97} \\ 
    \noalign{\vskip 0.8mm} %
    \cdashline{2-7}
    \noalign{\vskip 1.3mm} %
    & PVG      &\checkmark  & \cellcolor{tabsecond}29.77 &                         - &                          - & \cellcolor{tabsecond}27.19 \\ 
    & EmerNeRF &\checkmark  &                      25.14 &                         - &                          - & 23.49 \\ 
    & S3Gaussian &\checkmark&                      27.44 &                         - &                          - & 22.92 \\ 
    & DeSiRe-GS  &\checkmark& \cellcolor{tabthird}28.76  & \cellcolor{tabsecond}0.873& \cellcolor{tabsecond}0.193 & \cellcolor{tabthird}26.26\\ 
    & {\modelname} (ours) &\checkmark& \cellcolor{tabfirst}30.83  & \cellcolor{tabfirst}0.900 & \cellcolor{tabfirst}0.160  & \cellcolor{tabfirst}29.20 \\

    \midrule

    \multirow{8}{*}{\rotatebox[origin=c]{90}{\shortstack[c]{\small AD-GS setting}}}
    & StreetGS &$\times$  &                                {33.97} &                               {0.926} &                                {0.227} &                             {28.50} \\
    & 4DGS     &$\times$  &                                {34.64} &                               {0.940} &                                {0.244} &                             {29.77} \\
    & SplatAD  &$\times$  &                                {34.24} &                               {0.925} &                                {0.246} &                             {29.68} \\ 
    & SplatAD (CasTrack)  &$\times$  &                           {32.52} &                               {0.924} &                                {0.241} &                             {25.31} \\ 
    \noalign{\vskip 0.8mm} %
    \cdashline{2-7}
    \noalign{\vskip 1.3mm} %
    & PVG      &\checkmark  &                      29.54 &                     0.895 &                      0.266 & 21.56 \\ 
    & EmerNeRF &\checkmark  &                      31.32 &                     0.881 &                      0.301 & 21.80 \\ 
    & Grid4D   &\checkmark  & \cellcolor{tabthird}32.19  & \cellcolor{tabfirst}0.921 & \cellcolor{tabthird}0.253  & \cellcolor{tabthird}22.77 \\ 
    & AD-GS     &\checkmark  & \cellcolor{tabsecond}33.91 & \cellcolor{tabfirst}0.927 & \cellcolor{tabfirst}0.228  & \cellcolor{tabsecond}27.41 \\ 
    & {\modelname} (ours) &\checkmark& \cellcolor{tabfirst}34.59  & \cellcolor{tabfirst}0.929 & \cellcolor{tabsecond}0.235 & \cellcolor{tabfirst}29.63\\

    \midrule

    \multirow{2}{*}{\rotatebox[origin=c]{90}{\shortstack[c]{\small CoDa}}}
    
    & CoDa-4DGS &\checkmark  & \cellcolor{tabsecond}{28.66}  & \cellcolor{tabsecond}{0.900} &  \cellcolor{tabfirst}{0.058} & - \\
    
    & {\modelname} (ours) &\checkmark  & \cellcolor{tabfirst}{30.50} & \cellcolor{tabfirst}{0.875}  & \cellcolor{tabsecond}{0.090} & {-} \\   

    \midrule

    \multirow{2}{*}{\rotatebox[origin=c]{90}{\shortstack[c]{\small SF}}}
    
    & SplatFlow &\checkmark  &                                \cellcolor{tabsecond}{28.71} & \cellcolor{tabsecond}{0.874} & \cellcolor{tabsecond}{0.239} & {-} \\
    
    & {\modelname} (ours) &\checkmark  &                                \cellcolor{tabfirst}{29.95} & \cellcolor{tabfirst}{0.879} & \cellcolor{tabfirst}{0.183} & {-} \\
    
    \bottomrule
    \end{tabular}
    }
    \label{tab:results_both_splits}                                                                                                                                  
\end{table}

\begin{table}[t]
    \small
    \centering
    \setlength{\tabcolsep}{5.6pt}
    \caption{NVS results for lidar point cloud rendering. Our method obtains similar lidar rendering performance to the annotation-based SplatAD.}
    \begin{tabular}{l cccc}
    \toprule
                     &  Depth $\downarrow$ & Intensity $\downarrow$ & Drop acc. $\uparrow$ & CD $\downarrow$ \\ \midrule
 SplatAD             & 0.01  & 0.055  & 87.3 & 0.98   \\
 {\modelname} (ours) & 0.01  & 0.056  & 87.5  & 1.24   \\
    \bottomrule                                                                                           
    \end{tabular}                                                                                                         
    \label{tab:results_lidar}                                                   
\end{table}
\label{sec:implementation}

{\modelname} is implemented in \texttt{neurad-studio}~\cite{neuradstudio}, built upon the rasterization framework from SplatAD~\cite{hess2025splatad, splatad}. For pre-processing, we use Grounded-SAM-2~\cite{groundedsam2} to generate instance masks and DINOv3~\cite{simeoni2025dinov3} for image features. We optimize {\modelname} for 30,000 iterations, following the hyperparameter settings in \cite{hess2025splatad} unless otherwise specified. All experiments were run on an NVIDIA A100 40GB GPU. See \cref{sec:implementation_details} for further details.
\begin{figure}[t]
    \centering
    \includegraphics[width=1\linewidth,trim={0cm 0cm 0cm 0cm},clip]{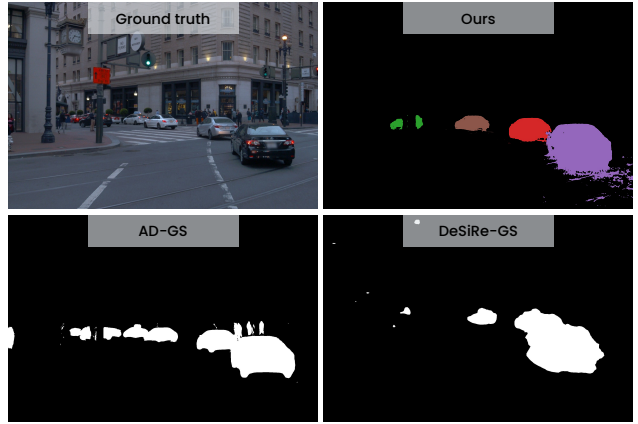}
    \caption{Dynamic mask rendering results. Beyond separating dynamic and static components, our method also renders instance masks for each dynamic object.}
    \label{fig:instance_masks}
\end{figure}
\subsection{Dataset}
\label{sec:datasets}
Our experiments are conducted on three subsets of Waymo Open Dataset. One is the set of eight sequences used in StreetGS \cite{yan2024streetgs} and AD-GS \cite{xu2025ad}, the second is the Waymo NeRF-On-The-Road (NOTR) dataset, a curated subset of challenging sequences provided in EmerNeRF \cite{yang2023emernerf}, and the last the set of 4 sequences used in PVG \cite{chen2026periodic}. NOTR contains 32 static, 32 dynamic, and 56 diverse sequences that cover various weather conditions and road types. %
See \cref{sec:dataset_details} for more details.

\begin{table}[t]
    \small
    \centering
    \setlength{\tabcolsep}{0.5pt}
    \caption{NVS results under varying view densities. Even with limited training views, our method achieves high DPSNR which highlights the effectiveness of our instance-decomposed model of dynamic objects. \colorbox{tabfirst}{First}, \colorbox{tabsecond}{second}, \colorbox{tabthird}{third}.}
    \resizebox{1.0\linewidth}{!}{
    \begin{tabular}{ll cc cc cc cc}
    \toprule
    &                       & \multicolumn{2}{c}{25\%} & \multicolumn{2}{c}{50\%} & \multicolumn{2}{c}{75\%} & \multicolumn{2}{c}{100\%} \\
    \cmidrule(lr){3-4} \cmidrule(lr){5-6} \cmidrule(lr){7-8} \cmidrule(lr){9-10}
    &                       & \small PSNR $\uparrow$  & \small DPSNR $\uparrow$     
                            & \small PSNR $\uparrow$  & \small DPSNR $\uparrow$    
                            & \small PSNR $\uparrow$  & \small DPSNR $\uparrow$    
                            & \small PSNR $\uparrow$  & \small DPSNR $\uparrow$ \\
    \midrule

    & DeSiRe-GS             & \cellcolor{tabthird}24.37 & \cellcolor{tabsecond} 22.97
                            & \cellcolor{tabthird}28.78 & \cellcolor{tabthird} 27.34
                            & \cellcolor{tabthird}30.04 &\cellcolor{tabthird}28.04
                            & \cellcolor{tabfirst}35.11 & \cellcolor{tabsecond}34.99\\

    & AD-GS                  & \cellcolor{tabsecond}26.21 & \cellcolor{tabthird}22.33
                            & \cellcolor{tabfirst}29.97 & \cellcolor{tabsecond}26.85
                            & \cellcolor{tabfirst}30.76 & \cellcolor{tabsecond}28.07
                            & \cellcolor{tabthird}34.42 & \cellcolor{tabfirst} 35.09\\

    & {\modelname} (ours)   & \cellcolor{tabfirst}26.83  &  \cellcolor{tabfirst}26.35                          
                            & \cellcolor{tabsecond}29.19  & \cellcolor{tabfirst}28.74                           
                            & \cellcolor{tabsecond} 30.11 & \cellcolor{tabfirst} 29.25                         
                            & \cellcolor{tabsecond} 35.04 &\cellcolor{tabthird}33.67 \\
    \bottomrule
    \end{tabular}
    }
    \label{tab:results_view_densities}                                                                                                                                  
\end{table}
\begin{figure*}[t]
    \centering
    \includegraphics[width=1\linewidth,trim={0cm 0cm 0cm 0cm},clip]{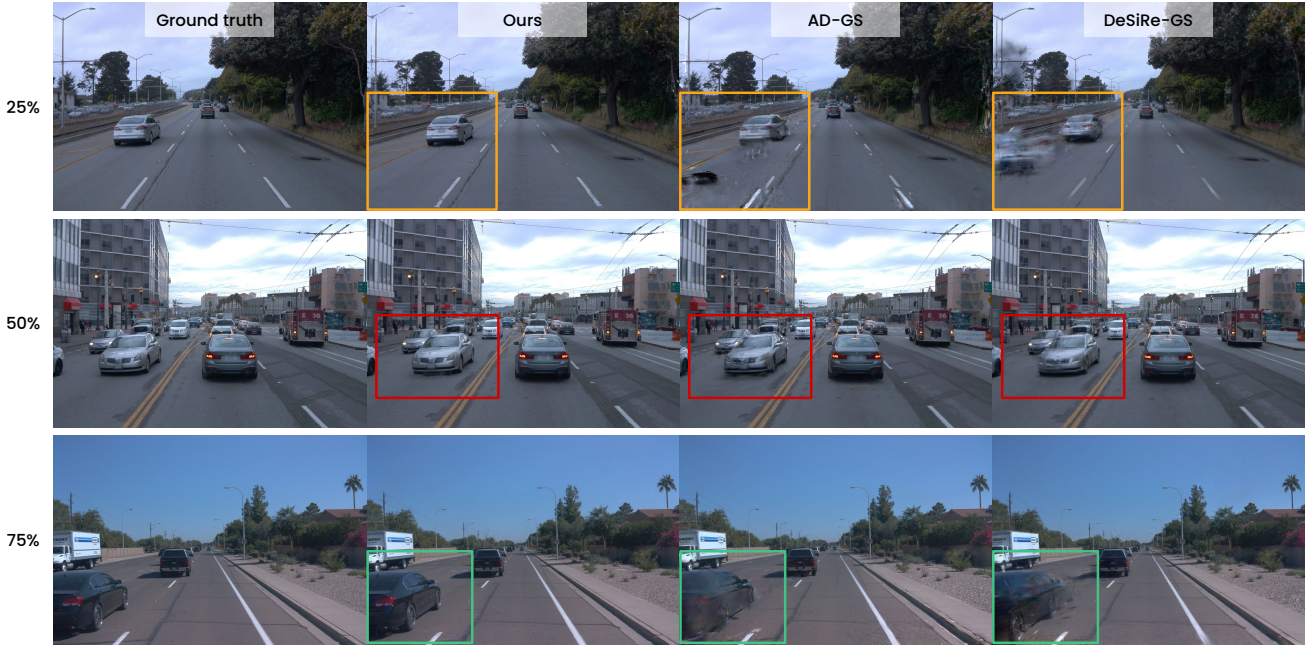}
    \caption{Qualitative comparisons of novel view synthesis over different view densities ($25\%$, $50\%$, and $75\%$ of training frames) on the dynamic subset of Waymo NOTR. Our instance-decomposed representation enables high-quality rendering of dynamic objects even when trained with sparse viewpoints.}
    \label{fig:qual_fig_collage}
\end{figure*}
\subsection{Baseline comparisons}
\label{sec:baselines}

We compare {\modelname} with state-of-the-art self-supervised rendering methods designed for dynamic automotive scenes, focusing our comparison with the following best performing methods, DeSiRe-GS \cite{peng2025desire}, AD-GS \cite{xu2025ad}, CoDa-4DGS \cite{song2025coda}, and SplatFlow \cite{sun2025splatflow}. In addition, we also report results for several supervised methods that rely on bounding box annotations to model dynamic objects. Among these, SplatAD \cite{splatad} is the most closely related to our approach and serves as a strong reference for assessing how closely {\modelname} approaches supervised performance. For further reference, we run SplatAD using tracks from CasTrack~\cite{wu2022castrack}, a high-performing tracker on the Waymo 3D tracking leaderboard.

\parsection{NVS results}
We report PSNR, SSIM \cite{zhou2004ssim}, and LPIPS \cite{zhang2018lpips} as our primary evaluation metrics, with LPIPS computed using the VGG network~\cite{simonyan2014very}. To specifically assess performance in dynamic regions, we also compute dynamic PSNR (DPSNR) using dynamic object masks. Since both {\modelname} and SplatAD support lidar rendering, we additionally evaluate lidar metrics for these methods, reporting median squared depth error, RMSE intensity error, ray drop accuracy, and chamfer distance (CD).

For a fair comparison with prior work, we adopt the data splits and evaluation protocols of each baseline. For DeSiRe-GS, we follow their setup of using 90\% of the frames for optimization and evaluating on every tenth frame. We conduct these experiments on the dynamic NOTR subset, using the three front cameras. Due to computational constraints of DeSiRe-GS, we use half-resolution images and only the first 50 frames in each sequence.
For AD-GS, we follow their protocol of using 75\% of the frames for optimization and evaluating on every fourth frame. Consistent with their settings, we use the eight sequences presented in StreetGS, the front camera, and full-resolution images. We use the same setting as DeSiRe-GS for CoDa-4DGS, but run the experiments on the complete NOTR dataset. To compare our results to SplatFlow, we use the same 4 Waymo sequences used in \cite{chen2026periodic} using 75\% of frames for training. To calculate DPSNR for each setting, we use the same dynamic masks originally used by each respective method as the ground truth.

We present comparisons with baseline methods in \cref{tab:results_both_splits}. {\modelname} achieves competitive or superior performance to prior work on both full-frame and dynamic-region evaluations. Notably, {\modelname} performs comparably to its supervised counterpart, SplatAD, and achieves similar performance on LiDAR metrics (\cref{tab:results_lidar}). {\modelname} further outperforms SplatAD when the latter uses tracks from CasTrack, demonstrating that our approach can outperform dataset-specific trackers.

\begin{figure}[t]
    \centering
    \includegraphics[width=1\linewidth,trim={0cm 0cm 0cm 0cm},clip]{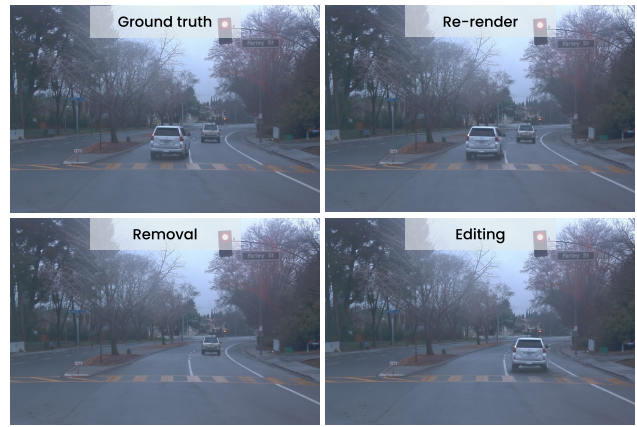}
    \caption{Instance editing. Our instance-decomposed representation enables targeted modifications of individual instances, such as their complete removal or the editing of their trajectories.}
    \label{fig:actor_editing}
\end{figure}

\parsection{Decomposition quality}
To better understand the differences in DPSNR across methods, \cref{fig:instance_masks} visualizes the dynamic object masks generated by DeSiRe-GS, AD-GS, and our method. AD-GS yields reasonably accurate dynamic regions, while DeSiRe-GS often over-segments static areas. Both methods only segment dynamic regions and cannot decompose individual object instances. In contrast, {\modelname} generates instance masks that closely align with ground truth. Since our model maintains a consistent set of Gaussians for each object, the appearance and geometry of each actor remain stable, leading to more coherent reconstructions and high rendering quality in dynamic regions. Further, \cref{fig:actor_editing} illustrates how our instance-decomposition enables targeted modification of objects. 

A side effect of our dynamic object model is that instance Gaussians may also represent nearby environmental elements that move with the object such as shadows or adjacent appearance effects. This can also be seen in \cref{fig:instance_masks}, where the orange vehicle mask slightly extends into the road surface.

\subsection{View density comparisons}
\label{sec:view_densities}
To assess the robustness of our approach, we conduct additional experiments under varying view densities. Specifically, we evaluate {\modelname} alongside DeSiRe-GS and AD-GS, using 25\%, 50\%, 75\%, and 100\% of the frames for optimization, while evaluating on the remaining frames, except for 100\% where all frames are also used for evaluation. The frames are selected linearly spaced. To enable a consistent comparison across methods, we adopt a unified setup, using the dynamic subset of NOTR with a single camera at full resolution and the first 50 frames of each sequence. We report PSNR and DPSNR %
in \cref{tab:results_view_densities}.

Both DeSiRe-GS and AD-GS perform well in the full reconstruction (100\%) setting, demonstrating the effectiveness of their time-varying parameterizations in fitting the data. However, evaluations under sparser settings reveal that these parameterizations struggle with larger interpolation gaps, as they are not explicitly constrained to capture the underlying dynamics. DeSiRe-GS degrades significantly on the sparser settings, especially in dynamic regions. AD-GS exhibits more stable performance across different frame densities, but also suffers from a drop in dynamic regions for more sparse settings. In contrast, the performance of {\modelname} in dynamic regions is much more consistent with the full-image results across all view densities, and surpasses the baselines with more than 4.8 PSNR in dynamic regions on the most sparse setting.

This is further illustrated in \cref{fig:qual_fig_collage}, which shows qualitative comparisons of novel view reconstruction on the dynamic subset of Waymo NOTR for different view densities. As also observed in quantitative results, {\modelname} maintains stable reconstruction quality and consistent geometry even when trained with fewer viewpoints. 

\subsection{Object class comparisons}
\label{sec:obj_class_comparisons}

For further analysis of our method, we present DPSNR for three different classes of road-users; vehicles, pedestrians, and cyclists. These experiments were run using the 50\% setting of the setup presented in \cref{sec:view_densities}. Masks of the dynamic regions were obtained by projected 3D bounding box annotations, and filtered based on speed and semantic class.
As shown in \cref{tab:class_psnrs}, {\modelname} attains the highest accuracy on vehicles, the only class that is modeled as dynamic instances in our approach. {\modelname} also shows strong results on pedestrians and cyclists, even though the rigid motion assumption is only partially valid for these classes.

\begin{table}[t]
    \small
    \centering
    \setlength{\tabcolsep}{1.2pt}
    \caption{NVS results filtered on different dynamic object classes. {\modelname} demonstrates strong performance for vehicles and pedestrians, with cyclists being more challenging. \colorbox{tabfirst}{First}, \colorbox{tabsecond}{second}, \colorbox{tabthird}{third}.}
    \begin{tabular}{ll ccccc}
    \toprule
    
    & & PSNR & $\text{DPSNR}_{\text{Vehicle}}$ & $\text{DPSNR}_{\text{Pedestrian}}$ & $\text{DPSNR}_{\text{Cyclist}}$ \\
    
    \midrule
    
    & DeSiRe-GS            & \cellcolor{tabthird}28.78 & \cellcolor{tabthird}25.04 & \cellcolor{tabsecond}27.65 & \cellcolor{tabfirst}29.34\\ 
    & AD-GS                 & \cellcolor{tabfirst}29.97 & \cellcolor{tabsecond}26.80 & \cellcolor{tabthird}27.22 & \cellcolor{tabthird}26.52 \\ 
    & {\modelname} (ours)  & \cellcolor{tabsecond}29.19 & \cellcolor{tabfirst}29.02 & \cellcolor{tabfirst}28.31 & \cellcolor{tabsecond}27.23 \\ 
    
    \bottomrule
    \end{tabular}
    \label{tab:class_psnrs}                                                                                                                                  
\end{table}

\subsection{Generalization}
To assess the generalization and robustness of our approach, we further evaluate it on PandaSet~\cite{pandaset}, without any hyperparameter tuning. Following \cite{splatad}, we use the same 10 sequences and all six available cameras at full resolution, and compute LPIPS using AlexNet~\cite{krizhevsky2012imagenet}. We perform novel view synthesis using 50\% of the frames for optimization and the remaining for evaluation, and compare our results against prior state-of-the-art methods. As shown in \cref{tab:results_pandaset}, {\modelname} achieves performance competitive with the best-performing method, SplatAD, despite not relying on any annotations.

\begin{table}[t]
    \small
    \centering
    \setlength{\tabcolsep}{5.8pt}
    \caption{NVS results on PandaSet, using all six cameras at full-resolution. {\modelname} achieves performance on par with annotation-based methods. }
    \begin{tabular}{l cccc}
    \toprule
                         &Anno. free&        PSNR $\uparrow$     &       SSIM $\uparrow$     &        LPIPS $\downarrow$  \\
    \midrule
    PVG                 &$\times$  &    24.01 &   0.712 &    0.452 \\
    Street-GS           &$\times$  &    24.73 &   0.745 &    0.314 \\   
    SplatAD             &$\times$  &    26.76 &   0.815 &    0.193 \\   
    \noalign{\vskip 0.8mm} %
    \cdashline{1-5}
    \noalign{\vskip 1.3mm} %
    {\modelname} (ours) &\checkmark&  26.78  &  0.814  & 0.174    \\ 
    \bottomrule
    \end{tabular}
    \label{tab:results_pandaset} 
\end{table}

\subsection{Ablations}
\label{sec:ablations}
We ablate the effectiveness of key components in our method by analyzing their impact on NVS performance, using the AD-GS setting described in \cref{sec:baselines} but with only 50\% of views used for optimization. The results of the ablations are presented in \cref{tab:ablations}. We observe that the initial filtering of lidar points using eroded masks (a) and DBSCAN (b) helps improve the final rendering results. Removing the RANSAC-based registration of points (c), and instead naively estimating pose translations by the point clouds centroids, has a severe impact on the performance. Selecting point correspondences using RGB colors instead of image features (d) also degrades the performance notably. Further, we see that the smoothing (e) is a key component for attaining good results in dynamic regions, while the outlier rejection from a first initial smoothing iteration (f) has a small impact on the final rendering results. Finally, we note the importance of refining the trajectories using gradients from the rendering losses during optimization (g).
Additionally, we analyze the impact of image features in \cref{tab:feature_ablations}. Specifically, we run the registration using different layers of DINOv3, image features from SAM2~\cite{ravi2024sam2}, and RGB colors. We observe that DINOv3 has the most descriptive features for successful registration, with earlier layers giving a slight increase in performance in dynamic regions.

\begin{table}[t]
    \small
    \centering
    \caption{NVS results when removing different model components.}
    \setlength{\tabcolsep}{3.8pt}
    \begin{tabular}{clcccc}
    \toprule
            &                   & PSNR $\uparrow$  & SSIM $\uparrow$ & LPIPS $\downarrow$ & DPSNR $\uparrow$  \\
          & Full model          & 33.59 & 0.920 & 0.240 & 29.35   \\ \midrule
       a) & Eroded masks        & 33.50 & 0.920 & 0.241 & 29.32   \\
       b) & DBSCAN              & 33.48 & 0.920 & 0.242 & 28.07   \\
       c) & Registration        & 32.60 & 0.915 & 0.247 & 26.67   \\
       d) & Image features      & 32.7 & 0.917 & 0.243 & 25.70   \\
       e) & Smoothing           & 32.96 & 0.918 & 0.244 & 28.01   \\
       f) & Outlier rejection   & 33.38 & 0.919 & 0.242 & 29.44   \\
       g) & Refinement          & 33.32 & 0.918 & 0.245 & 27.36   \\

    \bottomrule
    \end{tabular}
    \label{tab:ablations}
\end{table}
\begin{table}[t]
    \small
    \centering
    \setlength{\tabcolsep}{5.0pt}
    \caption{NVS results from using image features from different models when establishing point correspondences. \colorbox{gray}{Gray} row marks the image features used in our method.}    
    \begin{tabular}{lcccc}
    \toprule
                            & PSNR $\uparrow$  & SSIM $\uparrow$ & LPIPS $\downarrow$ & DPSNR $\uparrow$  \\
    \midrule
        RGB                 & 32.76 & 0.917 & 0.243 & 25.72   \\
        SAM2                & 33.24 & 0.918 & 0.245 & 27.74   \\
        \cellcolor{gray}DINOv3 layer 7      & \cellcolor{gray}33.59 & \cellcolor{gray}0.920 & \cellcolor{gray}0.240 & \cellcolor{gray}29.35   \\
        DINOv3 layer 9      & 33.57 & 0.919 & 0.240 & 29.17   \\
        DINOv3 layer 12     & 33.52 & 0.920 & 0.242 & 28.95   \\
    \bottomrule
    \end{tabular}
    \label{tab:feature_ablations}
\end{table}

\subsection{Tracking}
\label{sec:tracking}
While {\modelname} primarily targets high-quality scene reconstruction and rendering, we also evaluate the quality of the resulting motion trajectories. Specifically, we compare our optimized trajectories against human-annotated ground truth trajectories on a combined set of the 32 sequences from the dynamic subset of NOTR and eight sequences from the AD-GS split. We report Multiple Object Tracking Accuracy (MOTA)~\cite{bernardin2008evaluating} and Multiple Object Tracking Precision (MOTP)~\cite{bernardin2008evaluating} across different distance thresholds, along with detailed metrics such as false positives, ID switches, recall, and precision. The results are provided in \cref{sec:supp_tracking_results} along with qualitative examples and evaluation details. Since none of the baseline methods perform instance decomposition, tracking evaluation is not applicable to them. Nevertheless, we include our tracking results to establish a reference point and to encourage future research in this direction. 

Our trajectories exhibit occasional ID switches, as {\modelname} does not explicitly handle identity association across frames. Additional errors arise from stationary objects or incorrectly classified masks generated by Grounded-SAM-2. Furthermore, this evaluation does not account for potential constant offsets between the predicted and ground-truth trajectories, which can occur even under perfect motion tracking, due to partial or incomplete point cloud representations of objects.

\subsection{Limitations}

While IDSplat provides strong reconstruction performance and enables instance-level decomposition, several limitations remain. First, our object initialization and trajectory estimation are dependent on lidar measurements. Therefore, objects that fall outside of the lidar field of view but are still visible in the cameras are excluded and considered a part of the static scene. Second, our framework assumes that dynamic actors behave as rigid bodies. This works well for vehicles but is less suitable for highly deformable classes such as pedestrians and cyclists, which can lead to reduced reconstruction quality for these categories. Third, because each dynamic object is represented by a single set of Gaussians, nearby environmental effects such as shadows or reflections, which occur consistently in the scene and move with the object, may be represented by the instance. Finally, the current system lacks explicit track management mechanisms, as we do not merge overlapping tracks or handle ID switches, which may result in duplicated or incomplete instances in challenging scenarios. Addressing these limitations presents a promising direction for future work.
\section{Conclusion}
\label{sec:conclusion}

We presented {\modelname}, a zero-shot framework for instance decomposition and neural rendering in dynamic driving scenes. 
Extensive experiments show that {\modelname} achieves competitive or superior performance compared to state-of-the-art self-supervised approaches, and even matches the performance of annotation-based baselines. By representing each actor as a rigid instance, the method establishes a clear separation not only between dynamic and static scene elements, but also between individual dynamic objects, enabling precise scene editing such as object repositioning or removal. 

\subsection*{Acknowledgements}
We thank Adam Lilja and William Ljungbergh for valueable feedback.
This work was partially supported by the Wallenberg AI, Autonomous Systems and Software Program (WASP) funded by the Knut and Alice Wallenberg Foundation. This work was also partially supported by Vinnova, the Swedish Innovation Agency. Computational resources were provided by NAISS at \href{https://www.nsc.liu.se/}{NSC Berzelius}, partially funded by the Swedish Research Council, grant agreement no. 2022-06725.
{
    \small
    \bibliographystyle{ieeenat_fullname}
    \bibliography{main}
}

\clearpage
\setcounter{page}{1}
\maketitlesupplementary
\appendix

In this supplementary material, we provide additional details of our implementation, datasets, and baselines, as well as extended quantitative and qualitative results. In \cref{sec:implementation_details} we describe the details of our implementation, including all hyperparameters and training settings. \cref{sec:dataset_details} outlines the dataset splits used in our experiments, while \cref{sec:baseline_details} provides details regarding the baselines included for comparison. In \cref{sec:supp_tracking_results}, we present further quantitative and qualitative evaluations of our tracking performance. \cref{sec:runtime_analysis} reports inference and training times, along with a preprocessing breakdown across the components of our method. Finally, \cref{sec:supp_qualitative_results} showcases additional qualitative examples of novel view synthesis for both our approach and the baselines.

\section{Implementation details}\label{sec:implementation_details}
\parsection{Scene representation}
We initialize Gaussians using up to a maximum of 5M lidar points, using the color from projecting the points into the temporally closest camera to initialize the base color of each Gaussian. We add 2M additional points with random colors and positions, where half are sampled uniformly within lidar range, and half are sampled linearly in inverse distance outside lidar range. Using the densification strategy in \cite{kheradmand2024mcmc}, we allow the representation to grow up to a maximum of 10M Gaussians. All Gaussians are initialized with 50\% opacity and scaled to 20\% of the average distance to their three nearest neighbors. We use a feature dimension of 13 for each Gaussian's corresponding feature vector, and a feature dimension of 8 for the sensor-embeddings. Following \cite{splatad}, we decode lidar intensity and ray drop probability using a small MLP of 2 layers and a hidden dimension of 32. Similarly, we use a small CNN for decoding the view-dependent effects when rendering images. This CNN is implemented using two residual blocks with a hidden dimension 32, kernel size 3, and a linear output layer.

\parsection{Instance decomposition}
To generate object masks, we query Grounded-SAM-2 for \texttt{car}, \texttt{truck}, \texttt{van}, \texttt{bus}, \texttt{train}, \texttt{human}, \texttt{cyclist}, \texttt{bicycle}, and \texttt{pedestrian} instances and track these masks across the sequence to assign consistent instance IDs. We prompt every frame to reduce frames with missing masks. Each mask is eroded by 3 pixels before projecting lidar points, as described in \cref{sec:instance_decomposition}. We only consider lidar points within 80 meters of the sensor. The DBSCAN-clustering in \cref{sec:instance_decomposition} is performed with a maximum neighborhood distance of 0.5 meters and a minimum of 10 points in a neighborhood to determine a core point. We select the largest cluster as the initial 3D representation for that instance. The DINOv3 features used in \cref{sec:point_registration} are taken from layer 7 (of the ViT-B/16 version), upsampled to the image size using bilinear interpolation, and associated to the points in the cluster by projection.

\parsection{Trajectory estimation}
For computational efficiency, we sub-sample both the source and target point clouds in our registration to a maximum of 5000 points, selected randomly. The pose between the point clouds is estimated from 100,000 RANSAC iterations. Each iteration samples three point correspondences, obtained from cosine-similarity using DINOv3 features. We only use matches with a similarity higher than $0.8$. We define registration fitness as the ratio of target points that are within 10 cm of a source point after transformation, and use a fitness threshold of $0.5$ to determine whether a registration is successful or not. 

\parsection{Trajectory smoothing} 
Object instances with total trajectory displacement below 1 meter are converted to the static representation. For all remaining instances, we refine their RANSAC-derived poses via an iterative smoothing process using a GTSAM factor graph. The states include poses, velocities, and curvatures, while the RANSAC estimates serve as measurements. As described in \cref{sec:trajectory_smoothing}, we additionally estimate a single rotation shared across all timestamps to align the axis-aligned measurements with the motion model (which assumes the local x-axis is forward). Measurement factors are implemented as absolute-pose factors with a Huber loss (threshold 1.0) and diagonal noise with standard deviations $[0.1,\, 0.1,\, 0.1]$ rad for rotation and $[0.2,\, 0.2,\, 0.2]$ m for translation. 
The motion model is defined as
\begin{align}
    &\theta = \kappa v \Delta t \\ 
    &\Delta x = \frac{ \text{sin}(\theta) }{ \kappa } \\
    &\Delta y = \frac{ 1 - \text{cos}(\theta) }{ \kappa } \\
    &\Delta z = 0.0 \\
    &\Delta R = R_z(\theta),
\end{align}
where
\begin{equation}
R_{z}(\theta) =
\begin{bmatrix}
\cos\theta & -\sin\theta & 0 \\
\sin\theta & \cos\theta  & 0 \\
0 & 0 & 1
\end{bmatrix}.
\end{equation}
Note that pose states are first rotated by the shared rotation before applying the motion-model prediction. Motion-model factors connect successive states, where the residual is defined as the deviation from the predicted motion. These factors use diagonal noise with standard deviations $[0.1,\, 0.1,\, 0.1]$ rad for rotation and $[0.2,\, 0.2,\, 0.2]$ m for translation. Velocity and curvature states follow random-walk priors with discretized standard deviations $\sqrt{0.5\Delta t}$ and $\sqrt{10^{-5}\Delta t}$, respectively. Pose states are further regularized to maintain moderate roll and pitch angles via magnitude-based residuals (computed after applying the shared rotation), with diagonal noise and standard deviation $0.4$ for both roll and pitch. A curvature prior with standard deviation of $0.01$ rad is also applied at every timestep.

We optimize using GTSAM´s Levenberg-Marquardt solver. We run a single iteration to identify outlier measurements whose whitened error exceeds $1.345$, and then run the final optimization without those measurements for a maximum of ten iterations. 

\parsection{Scene optimization}
We jointly optimize all model and pose parameters for 30{,}000 steps with the Adam optimizer~\cite{kingma2014adam}. Learning rates and scheduling parameters are reported in \cref{tab:lr_table}. Following \cite{kerbl20233Dgaussians}, we adopt a resolution-scheduling scheme in which the first 3{,}000 optimization steps use images downsampled by a factor of 4, the next 3{,}000 steps use a downsampling factor of 2, and the remaining steps are performed with the original image size. 
\begin{table}
    \centering
    \caption{Learning rates (LR) for each parameter group. Exponential decay is used for scheduling when applicable.}
    \resizebox{0.99\linewidth}{!}{
    \begin{tabular}{l r r r}
    \toprule
    Parameters & Initial LR & Final LR & Warm-up steps \\
    \hline
Gaussian means & 1.6e-4 & 1.6e-6 & 0 \\
Gaussian opacities & 5.0e-2 & 5.0e-2 & 0 \\
Gaussian scales & 5.0e-3 & 5.0e-3 & 0 \\
Gaussian quaternions & 1.0e-3 & 1.0e-3 & 0 \\
Gaussian features & 2.5e-3 & 2.5e-3 & 0 \\
Decoders & 1.0e-3 & 1.0e-3 & 500 \\
Dynamic object positions  & 1.0e-3 & 1.0e-4 & 1500 \\
Dynamic object rotations  & 5.0e-5 & 1.0e-6 & 1500 \\
Sensor embeddings  & 1.0e-3 & 1.0e-3 & 500 \\
Sensor vel. linear  & 1.0e-3 & 1.0e-6 & 1000 \\
Sensor vel. angular  & 2.0e-4 & 1.0e-7 & 1000 \\
Cam. time to center  & 2.0e-4 & 1.0e-7 & 10000 \\
    \bottomrule
    \end{tabular}
    }
    \label{tab:lr_table}
\end{table}

We use the loss formulation and hyperparameter settings from \cite{hess2025splatad}, without additional tuning. While these baseline values proved robust, we note that dataset-specific tuning may yield further improvements. As in \cite{hess2025splatad}, we employ the MCMC described in \cref{eq:loss_fn} and adapted from \cite{kheradmand2024mcmc}, which include opacity and scale regularization:
\begin{equation}
    \lambda_{\text{MCMC}} \mathcal{L}_{\text{MCMC}} = \lambda_o \sum_i  \left| o_i \right| + \lambda_\Sigma  \sum_{ij} \left| \sqrt{ \text{eig}_j (\Sigma_i) } \right|.
\end{equation}
All loss-related hyperparameters are reported in \cref{tab:loss_table}.
\begin{table}
    \centering
    \caption{Hyperparameters used for loss weighting.}
    \begin{tabular}{l r}
    \toprule
    Loss parameter          & Weight \\
    \midrule
$\lambda_r$                 & 0.8\\
$\lambda_\text{depth}$      & 0.1\\
$\lambda_\text{los}$        & 0.1\\
$\lambda_\text{inten}$      & 1.0\\
$\lambda_\text{raydrop}$    & 0.1\\
$\lambda_o$                 & 5e-3\\
$\lambda_\Sigma$            & 1e-3\\
    \bottomrule
    \end{tabular}
    \label{tab:loss_table}
\end{table}

\section{Dataset details}\label{sec:dataset_details}

In this section, we provide additional details about the datasets used in our experiments. We evaluate on three subsets of the Waymo Open Dataset, chosen to match the baseline settings, and include PandaSet to demonstrate the robustness of our method. 

\parsection{StreetGS split} This subset of Waymo, used for the AD-GS evaluation setting, contains 8 sequences of roughly 100 frames, each featuring a variety of moving vehicles. These sequences do not include pedestrians or cyclists. We followed instructions in the official implementation of StreetGS \cite{yan2024streetgs} to preprocess this subset. The original segments used to construct this split, along with their start and end frames, are listed in \cref{tab:streetgs_segments}.

\begin{table*}[t]
    \small
    \centering
    \setlength{\tabcolsep}{5.6pt}
    \caption{StreetGS sequences and the corresponding Waymo Open Dataset segments.}
    \begin{tabular}{l lrr}
    \toprule
                     Sequence ID & Segment Name & Start Frame & End Frame \\ \midrule
                     006&  segment-10448102132863604198\_472\_000\_492\_000 & 0  & 85   \\
                     026& segment-12374656037744638388\_1412\_711\_1432\_711  & 0  & 100   \\
                     090& segment-17612470202990834368\_2800\_000\_2820\_000  & 0  & 102   \\
                     105& segment-1906113358876584689\_1359\_560\_1379\_560  & 20  & 186   \\
                     108& segment-2094681306939952000\_2972\_300\_2992\_300  & 20  & 115   \\
                     134& segment-4246537812751004276\_1560\_000\_1580\_000  & 106  & 198   \\
                     150& segment-5372281728627437618\_2005\_000\_2025\_000  & 96  & 197   \\
                     181& segment-8398516118967750070\_3958\_000\_3978\_000  & 0  & 160   \\
    \bottomrule                                                                                           
    \end{tabular}                                                                      
    \label{tab:streetgs_segments}                                                   
\end{table*}

\parsection{NOTR} NeRF-On-The-Road (NOTR) is a curated set of 120 driving sequences from Waymo spanning a broad range of challenging conditions. It includes 32 \textit{Static} scenes, 32 \textit{Dynamic} scenes with multiple moving actors, and 56 \textit{Diverse} scenes capturing variations in driving speed, weather, and lighting conditions. We preprocessed this data following the official code of \cite{yang2023emernerf}. \cref{tab:dynamic_notr_segments} lists all Waymo data segments included in this subset. We use \textit{Dynamic} NOTR sequences for the DeSiRe-GS setting and the complete NOTR dataset for the CoDa-4DGS setting.

\begin{table*}[t]
    \small
    \centering
    \setlength{\tabcolsep}{5.6pt}
    \caption{Dynamic NOTR sequences and the corresponding Waymo Open Dataset segments (-1 denotes the last frame).}
    \begin{tabular}{l lrr}
    \toprule
                     Sequence ID & Segment Name & Start Frame & End Frame \\ \midrule
                    016 & segment-10231929575853664160\_1160\_000\_1180\_000 &0 & -1 \\
                    021 & segment-10391312872392849784\_4099\_400\_4119\_400 &0 & -1 \\
                    022 & segment-10444454289801298640\_4360\_000\_4380\_000 &0 & -1 \\
                    025 & segment-10498013744573185290\_1240\_000\_1260\_000 &0 & -1 \\
                    031 & segment-10588771936253546636\_2300\_000\_2320\_000 &0 & -1 \\
                    034 & segment-10625026498155904401\_200\_000\_220\_000 &0 & -1 \\
                    035 & segment-10664823084372323928\_4360\_000\_4380\_000 &0 & -1 \\
                    049 & segment-10963653239323173269\_1924\_000\_1944\_000 &0 & -1 \\
                    053 & segment-11017034898130016754\_697\_830\_717\_830 &0 & -1 \\
                    080 & segment-11718898130355901268\_2300\_000\_2320\_000 &0 & -1 \\
                    084 & segment-11846396154240966170\_3540\_000\_3560\_000 &0 & -1 \\
                    086 & segment-1191788760630624072\_3880\_000\_3900\_000 &0 & -1 \\
                    089 & segment-11928449532664718059\_1200\_000\_1220\_000 &0 & -1 \\
                    094 & segment-12027892938363296829\_4086\_280\_4106\_280 &0 & -1 \\
                    096 & segment-12161824480686739258\_1813\_380\_1833\_380 &0 & -1 \\
                    102 & segment-12251442326766052580\_1840\_000\_1860\_000 &0 & -1 \\
                    111 & segment-12339284075576056695\_1920\_000\_1940\_000 &0 & -1 \\
                    222 & segment-14810689888487451189\_720\_000\_740\_000 &0 & -1 \\
                    323 & segment-16801666784196221098\_2480\_000\_2500\_000 &0 & -1 \\
                    382 & segment-18111897798871103675\_320\_000\_340\_000 &0 & -1 \\
                    402 & segment-1918764220984209654\_5680\_000\_5700\_000 &0 & -1 \\
                    427 & segment-2259324582958830057\_3767\_030\_3787\_030 &0 & -1 \\
                    438 & segment-2547899409721197155\_1380\_000\_1400\_000 &0 & -1 \\
                    546 & segment-4414235478445376689\_2020\_000\_2040\_000 &0 & -1 \\
                    581 & segment-5083516879091912247\_3600\_000\_3620\_000 &0 & -1 \\
                    592 & segment-5222336716599194110\_8940\_000\_8960\_000 &0 & -1 \\
                    620 & segment-5835049423600303130\_180\_000\_200\_000 &0 & -1 \\
                    640 & segment-6242822583398487496\_73\_000\_93\_000 &0 & -1 \\
                    700 & segment-7670103006580549715\_360\_000\_380\_000 &0 & -1 \\
                    754 & segment-8822503619482926605\_1080\_000\_1100\_000 &0 & -1 \\
                    795 & segment-9907794657177651763\_1126\_570\_1146\_570 &0 & -1 \\
                    796 & segment-990914685337955114\_980\_000\_1000\_000 &0 & -1 \\
    \bottomrule                                                                                           
    \end{tabular}                                                                      
    \label{tab:dynamic_notr_segments}                                                   
\end{table*}

\parsection{PVG split} This set of 4 sequences from Waymo were used in \cite{chen2026periodic} and adopted by SplatFlow in their experiments. Further details of these sequences is presented in \cref{tab:pvg_segments}.

\begin{table*}[t]
    \small
    \centering
    \setlength{\tabcolsep}{5.6pt}
    \caption{PVG sequences and the corresponding Waymo Open Dataset segments.}
    \begin{tabular}{l lrr}
    \toprule
                     Sequence ID & Segment Name & Start Frame & End Frame \\ \midrule
                     017&  segment-10235335145367115211\_5420\_000\_5440\_000 & 61  & 109   \\
                     022& segment-13186511704021307558\_2000\_000\_2020\_000  & 26  & 74   \\
                     050& segment-13207915841618107559\_2980\_000\_3000\_000  & 6  & 54   \\
                     081& segment-13506499849906169066\_120\_000\_140\_000  & 26  & 74   \\
    \bottomrule                                                                                           
    \end{tabular}                                                                      
    \label{tab:pvg_segments}                                                   
\end{table*}

\parsection{Pandaset} PandaSet is a multimodal autonomous driving dataset containing camera and lidar data captured in diverse urban environments. We used the following 10 sequences from PandaSet: \texttt{001}, \texttt{011}, \texttt{016}, \texttt{028}, \texttt{053}, \texttt{063}, \texttt{084}, \texttt{106}, \texttt{123}, and \texttt{158}. 

Across our experiments, we selected the appropriate data based on the baselines under comparison and the corresponding cameras, image resolution and train-test splits. Full details are given in \cref{tab:experiment_settings}.

\begin{table*}[t]
    \small
    \centering
    \setlength{\tabcolsep}{5.6pt}
    \caption{Evaluation settings for tables in the main paper. * denotes that we use the full segments from StreetGS instead of using the start and end frames reported in \cref{tab:streetgs_segments}.}
    \begin{tabular}{l lcccc}
    \toprule
                     Table & Data & Num. Sequences & Cameras & Image Res. & Train. views \\ \midrule
                     
                     \cref{tab:results_both_splits} (DeSiRe-GS)& Dynamic NOTR & 32 & front, front\_left, front\_right & $[640\times960]$ & 90\% \\
                     
                     \cref{tab:results_both_splits} (AD-GS)& StreetGS Split & 8 & front & $[1280\times1920]$ & 75\% \\

                     \cref{tab:results_both_splits} (CoDa)& StreetGS Split & 120 & front, front\_left, front\_right & $[640\times960]$ & 90\% \\

                     \cref{tab:results_both_splits} (SF)& PVG Split & 4 & front, front\_left, front\_right & $[640\times960]$ & 75\% \\
                     
                     \cref{tab:results_lidar}& Dynamic NOTR & 32 & front, front\_left, front\_right & $[640\times960]$ & 90\% \\
                     
                     \cref{tab:results_view_densities}& Dynamic NOTR & 32 & front & $[1280\times1920]$ & 25\%, 50\%, 75\% \\
                     
                     \cref{tab:class_psnrs}& Dynamic NOTR & 32 & front & $[1280\times1920]$ & 50\% \\
                     
                     \cref{tab:results_pandaset}& PandaSet & 10 & all 6 cameras & $[1920\times1080]$ & 50\% \\
                     
                     \cref{tab:ablations}& StreetGS Split$^*$ & 8 & front & $[1280\times1920]$ & 50\% \\
                     
                     \cref{tab:feature_ablations}& StreetGS Split$^*$ & 8 & front & $[1280\times1920]$ & 50\% \\
    \bottomrule                                                                                           
    \end{tabular}                                                                      
    \label{tab:experiment_settings}                                                   
\end{table*}

\parsection{Dynamic ground truth masks} Both the StreetGS split and NOTR provide preprocessed dynamic masks obtained by projecting Waymo's 3D bounding box annotations onto the image plane and filtering objects based on their speed. These masks, however, are binary segmentation masks and do not include object class information. To evaluate the DPSNR over different classes of road-users, we generate our own 2D dynamic object masks from the same 3D bounding boxes using the same procedure. We create 3 sets of masks for vehicles, pedestrians, and cyclists. In addition, we apply an extra filter: bounding boxes with fewer than 10 associated lidar points are discarded to ensure that evaluation only considers objects that are observed by lidar. In \cref{tab:results_both_splits} and \cref{tab:results_lidar}, we use the class-agnostic dynamic masks provided by each dataset for computing DPSNR. In all other tables, we employ our own generated dynamic masks. Unless stated otherwise, DPSNR is computed using ground truth masks for moving vehicles. 

\section{Baseline details}\label{sec:baseline_details}

To obtain results for our two main baselines, DeSiRe-GS and AD-GS, we use their official implementation for training, evaluation, and visualization. We only modify configuration parameters for camera selection, image resolution, and train-test data split to match each experiment setting. All other hyperparameters remain the same, and the training follows the schedules reported in the original papers. For both methods, we adapt the official rendering scripts to extract the dynamic masks shown in \cref{fig:instance_masks} from their rendered output. 

Despite using the official code and training parameters, we could not reproduce the DeSiRe-GS results that they reported in \textit{Table 4} in their paper. We therefore report both the paper's numbers and our reproduced results in \cref{tab:desiregs_paper} for transparency. 

\begin{table}[t]
    \small
    \centering
    \setlength{\tabcolsep}{5.0pt}
    \caption{DeSiRe-GS NVS results for Dynamic NOTR with 90\% training views. SSIM and LPIPS were not reported in the paper.} 
    \begin{tabular}{lcc}
    \toprule
                            & PSNR $\uparrow$  & DPSNR $\uparrow$  \\
    \midrule
    Reported results (Tab. 4 in \cite{peng2025desire})                  & 30.45 & 28.66   \\
    Reproduced results (\cref{tab:results_both_splits})             & 28.76 &  26.26  \\
    \bottomrule
    \end{tabular}
    \label{tab:desiregs_paper}
\end{table}

\section{Tracking results}\label{sec:supp_tracking_results}
\begin{figure*}[t]
    \centering
    \includegraphics[width=1\linewidth,trim={0cm 0cm 0cm 0cm},clip]{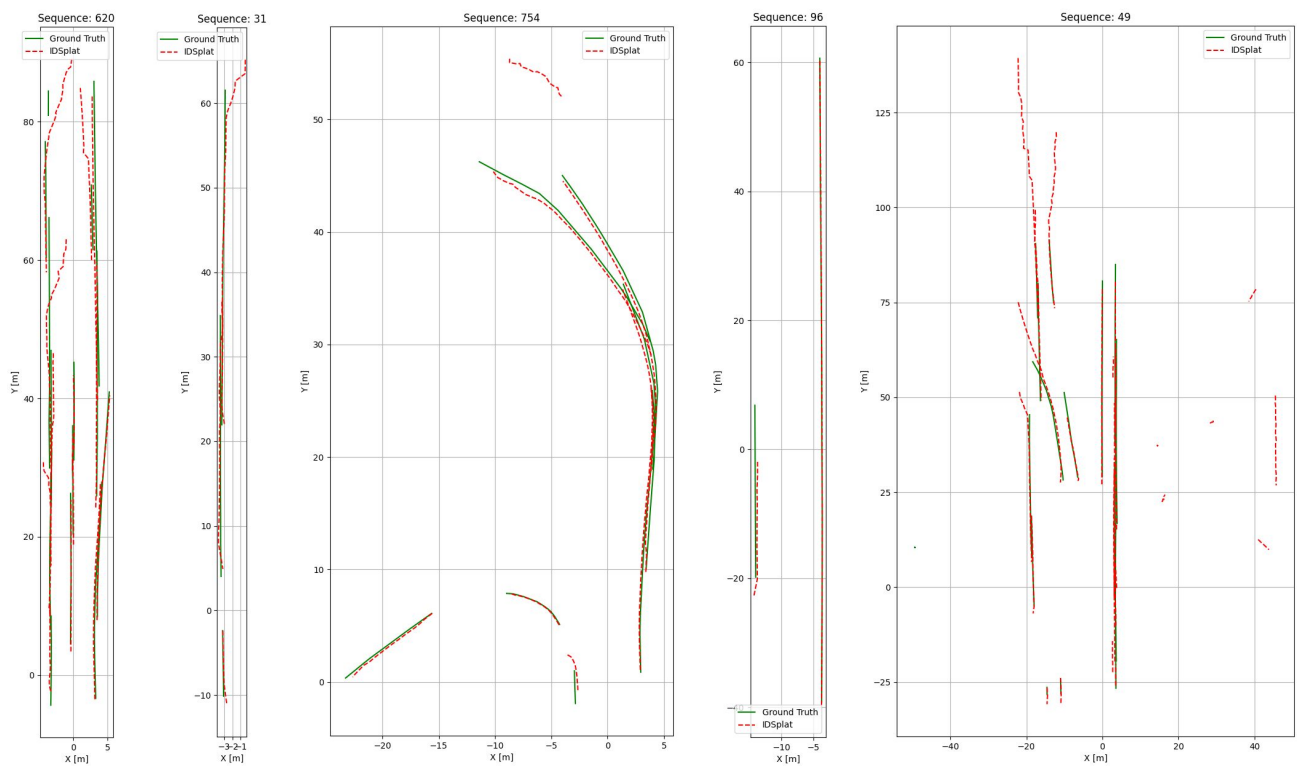}
    \caption{Plots of optimized trajectories from {\modelname} compared to ground-truth, for five different sequences.}
    \label{fig:tracking_fig}
\end{figure*}

We report the tracking performance of {\modelname} in \cref{tab:results_tracking}, evaluated over the combined 40 sequences from the NOTR dynamic subset and the AD-GS split. We provide Multiple Object Tracking Accuracy (MOTA)~\cite{bernardin2008evaluating} and Multiple Object Tracking Precision (MOTP)~\cite{bernardin2008evaluating}, along with the full set of underlying metrics used to compute them. \textit{Frames} corresponds to the total number of processed frames, while \textit{Objects} and \textit{Predictions} represent the total number of ground-truth and predicted object appearances, respectively (i.e., not counts of unique object identities). \textit{Matches} refers to the number of ground-truth-prediciton pairs that fall inside the distance threshold and are assigned via the Hungarian algorithm. \textit{Switches} counts the number of cases where a ground-truth identity is associated to different predicted identities over time. \textit{FP} denotes false positives (predictions with no corresponding ground-truth object), and \textit{FN} denotes false negatives (ground-truth objects with no corresponding prediction). MOTA is derived from these as
\begin{equation}
    \text{MOTA} = 1 - \frac{\text{FP} + \text{FN} + \text{Switches}}{\text{Objects}},
\end{equation}
while MOTP measures the localization error for matched pairs, averaged over all matches. \textit{Recall} quantifies the fraction of ground-truth objects that were detected, and \textit{Precision} quantifies the fraction of correct predictions. To provide a comprehensive view of performance, we compute these metrics over six distance thresholds: $0.5$ m, $1.0$ m, $2.0$ m, $3.0$ m, $5.0$ m and $10.0$ m. Qualitative examples are shown in \cref{fig:tracking_fig}.
\begin{table*}[t]
    \small
    \centering
    \caption{Tracking performance of {\modelname} on the combined 40 sequences from the NOTR dynamic subset and the AD-GS split, evaluated over different distance thresholds for the matching. MOTA and MOTP denotes Multiple Object Tracking Accuracy~\cite{bernardin2008evaluating} respectively Multiple Object Tracking Precision~\cite{bernardin2008evaluating}.}
    \resizebox{1.0\linewidth}{!}{
    \begin{tabular}{ccccccccccccc}
    \toprule
      Dist. threshold [m]      & \# Frames & \# Objects & \# Predictions & \# Matches & \# Switches $\downarrow$ & \# FP $\downarrow$ & \# FN $\downarrow$ & MOTA $\uparrow$ & MOTP $\downarrow$ & Recall $\uparrow$ & Precision $\uparrow$ \\
      \midrule
      $0.5$  & 2024 & 10295 & 11160 & 2769 & 83 & 8308 & 7443 & -0.54 & 0.27 & 0.28 & 0.26  \\
      $1.0$  & 2024 & 10295 & 11160 & 3835 & 119 & 7206 & 6341 & -0.33 & 0.41 & 0.38 & 0.35 \\
      $2.0$  & 2024 & 10295 & 11160 & 6487 & 203 & 4470 & 3605 & 0.20 & 0.87 & 0.65 & 0.60 \\
      $3.0$  & 2024 & 10295 & 11160 & 7528 & 223 & 3409 & 2544 & 0.40 & 1.07 & 0.75 & 0.69 \\
      $5.0$  & 2024 & 10295 & 11160 & 7717 & 232 & 3211 & 2346 & 0.44 & 1.16 & 0.77 & 0.71 \\ 
      $10.0$ & 2024 & 10295 & 11160 & 7903 & 270 & 2987 & 2122 & 0.48 & 1.71 & 0.79 & 0.73 \\
    \bottomrule
    \end{tabular}
    }
    \label{tab:results_tracking}                                                                                                                                  
\end{table*}

As expected, increasing the distance threshold yields more matches, reducing both the number of false positives and false negatives and thereby improving MOTA, precision, and recall. However, this comes at the cost of more identity switches and reduced localization accuracy (higher MOTP error). The low number of matches at smaller thresholds can partly be attributed to constant offsets between the predicted and ground-truth trajectories, which may arise even under perfect motion tracking due to partial or incomplete point cloud representations of predicted objects. Nevertheless, the results also highlight several opportunities for future work, including inter-frame identity association and more advanced strategies for modeling object births and deaths.

\section{Runtime analysis}\label{sec:runtime_analysis}
\cref{tab:inference_train_speed} reports the median inference rate and median total training time for our method compared to DeSiRe-GS and AD-GS, each evaluated in their respective setting using official implementations. \cref{tab:preprocess_breakdown} further breaks down preprocessing time across the components of {\modelname}. The preprocessing time is included in the total training times reported above.

\begin{table}
    \small
    \centering
    \resizebox{\linewidth}{!}{
    \begin{tabular}{l cc}
         & Inference rate [MP/s] & Train time [min] \\
         \midrule
         DeSiRe-GS & 0.9 & 413.1 \\
         IDSplat & \textbf{29.6} & \textbf{63.7} \\
         \midrule
         AD-GS & 12.2 & 150.6 \\
         IDSplat & \textbf{51.5} & \textbf{116.5} \\
    \end{tabular}
    }
    \caption{Inference rate in megapixels per second (MP/s) and total training time in minutes, compared against DeSiRe-GS and AD-GS in their respective settings.}
    \label{tab:inference_train_speed}
\end{table}

\begin{table}[]
    \small
    \centering
    \resizebox{1.0\linewidth}{!}{
    \begin{tabular}{l ccccc}
         Exp. setting & Tot. time [s] & Mask gen. & Point paint. & Registration & Smoothing \\
         \hline
         AD-GS & 127.6 & 79.8 (62.5\%) & 12.7 (9.9\%) & 33.5 (26.3\%) & 1.6 (1.3\%) \\
         DeSiRe & 98.9 & 50.1 (50.6\%) & 26.4 (26.7\%) & 20.3 (20.5\%) & 2.1 (2.2\%) \\
    \end{tabular}
    }
    \caption{Per-component preprocessing time breakdown for {\modelname}, reported in seconds with percentage of total. This preprocessing time is included in the total training times of \cref{tab:inference_train_speed}.}
    \label{tab:preprocess_breakdown}
\end{table}

\section{Qualitative results}\label{sec:supp_qualitative_results}
\begin{figure*}[t]
    \centering
    \includegraphics[width=1\linewidth,trim={0cm 0cm 0cm 0cm},clip]{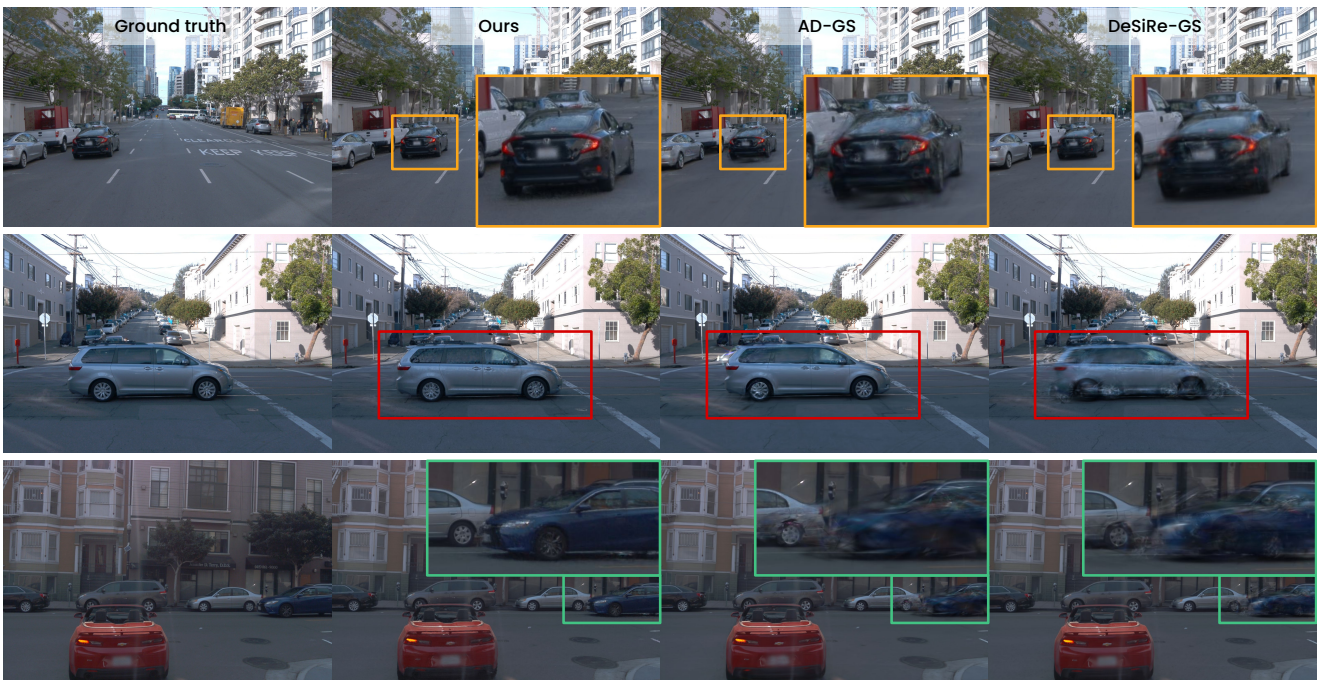}
    \caption{Qualitative comparison with baselines on the dynamic subset of Waymo NOTR, using 75\% of views for training.}
    \label{fig:suppmat_qualitatives}
\end{figure*}

We provide additional qualitative examples in \cref{fig:suppmat_qualitatives}, depicting NVS results for {\modelname}, AD-GS and DeSiRe when using 75\% of the views for training. All examples show validation views, and all sequences are from the dynamic subset of NOTR.

Additional qualitative results on deformable object classes are shown in \cref{fig:limitations}. Despite rigid-body modeling, pedestrians and cyclists are rendered with high fidelity. Rigid motion captures the dominant movement while small deformations are absorbed by view-dependent effects.

\begin{figure*}[t]
    \centering
    \includegraphics[width=1\linewidth,trim={0cm 0cm 0cm 0cm},clip]{figures/limitations_fig_v2.pdf}
    \caption{Qualitative results on deformable object classes. Pedestrians and cyclists are rendered with high fidelity despite being modeled as rigid.}
    \label{fig:limitations}
\end{figure*}

\immediate\closein\imgstream

\end{document}